\documentclass[conference]{IEEEtran}
\IEEEoverridecommandlockouts
\renewcommand{\baselinestretch}{0.96}
\usepackage{cite}
\usepackage{amsmath,amssymb,amsfonts}
\usepackage{algorithmic}
\usepackage{graphicx}
\usepackage{textcomp}
\usepackage{xcolor}

\usepackage{adjustbox}

\usepackage{hhline}

\def\BibTeX{{\rm B\kern-.05em{\sc i\kern-.025em b}\kern-.08em
    T\kern-.1667em\lower.7ex\hbox{E}\kern-.125emX}}
\begin{document}

\title{An Explainable Model-Agnostic Algorithm for CNN-based Biometrics Verification
}

\author{\IEEEauthorblockN{Fernando Alonso-Fernandez$^*$, Kevin Hernandez-Diaz$^*$, José M. Buades$^\dagger$, Prayag Tiwari$^*$, Josef Bigun$^*$}
\IEEEauthorblockA{$^*$ Halmstad University. Box 823. SE 301-18 Halmstad, Sweden\\
$^\dagger$ Computer Graphics and Vision and AI Group, University of Balearic Islands, Spain\\
Emails:  feralo@hh.se, kevin.hernandez-diaz@hh.se, josemaria.buades@uib.es, prayag.tiwari@hh.se, josef.bigun@hh.se}}

\maketitle

\begin{abstract}
This paper describes an adaptation of the Local Interpretable Model-Agnostic Explanations (LIME) AI method to operate under a biometric verification setting. LIME was initially proposed for networks with the same output classes used for training, and it employs the softmax probability to determine which regions of the image contribute the most to classification. However, in a verification setting, the classes to be recognized have not been seen during training. In addition, instead of using the softmax output, face descriptors are usually obtained from a layer before the classification layer. The model is adapted to achieve explainability via cosine similarity between feature vectors of perturbated versions of the input image.
The method is showcased for face biometrics with two CNN models based on MobileNetv2 and ResNet50.
\end{abstract}


\section{Introduction}
\label{sec:intro}

Biometric applications have a wide range of impacts on people's daily lives.
They cover scenarios such as identity verification for physical and digital access, border control, e-commerce, e-education, device unlocking, watchlist surveillance, forensic analysis, and more.
With the advent of deep learning methods \cite{[Sundararajan18-DLbiometrics]}, concerns have arisen regarding their explainability and interpretability.
Their black box nature raises questions about why a recognition system makes certain decisions and which parts of the input image are used \cite{Jain22PAMI_biometrics_trust_verify}.

In this research, we propose the use of LIME (Local Interpretable Model-agnostic Explanation) \cite{Ribeiro16LIME} for biometric verification, with a specific focus on face recognition (FR). However, our approach can be applied to other modalities too.
LIME was initially developed to provide explanations for classifiers by highlighting the pixels that contribute more positively to a specific class (in our case, a particular identity).
%
%
However, in biometric verification, the network is not used to classify the input into a predefined set of classes, but rather to generate a feature vector representing the input image.
This is done by removing the classification head and using the output of previous layers (typically the last one, although earlier layers can also be utilized) \cite{[Hernandez18]}.
Consequently, two feature vectors can be compared via distance metrics to produce a score that indicates the similarity between the images.
In this operational mode, the training classes (identities) are not necessarily the same used for verification.
Indeed, the latter may not be possible when a system is deployed to third parties.

%
%
%
%
%
%

%
%
%
%
%

\begin{figure}[b]
\centerline{\includegraphics[width=0.48\textwidth]{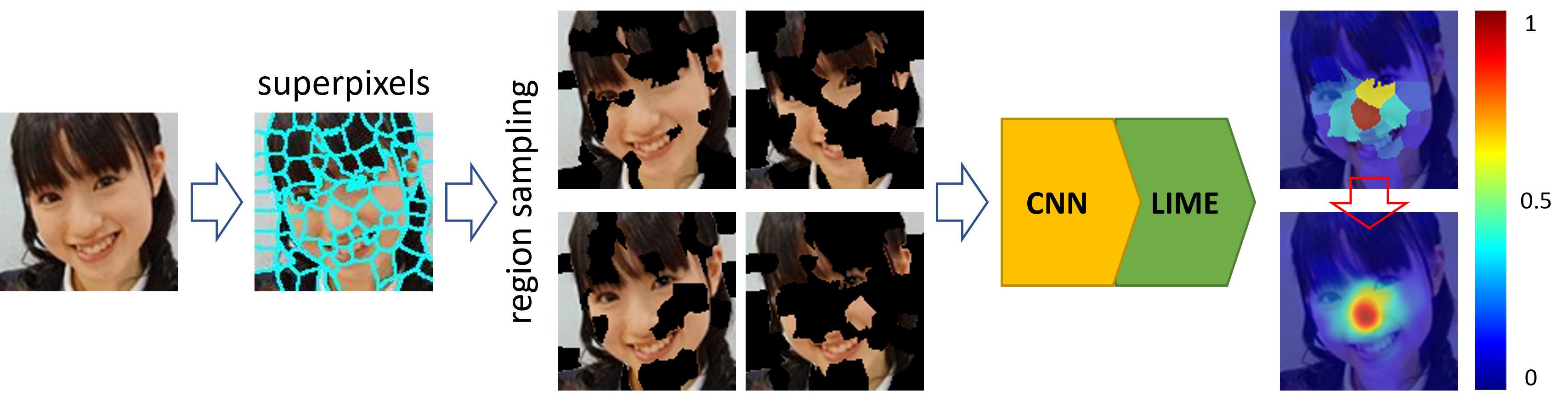}}
\caption{Overview of the explainability method.}
\label{fig:LIME_method}
\end{figure}

Several works aim to interpret face recognition (FR) models, but only a few address the importance of image regions.
One approach \cite{Yin19_ICCV_Interpretable_FR} constrains learning so that features directly relates to different face areas (measured by saliency maps), but it requires re-training, preventing to use FR models out-of-the-box.
%
%
In \cite{Dhar20FG_Attributes_in_Face_CNN}, correlations between attributes (age, gender, and pose) and CNN feature vectors are explored, enabling attribute inference from deep FR representations.
%
%
Estimating feature uncertainty as a measure of quality was studied \cite{ShiJain19ICCV_ProbabilisticFaceEmbeddings} by representing each image as a Gaussian distribution in the latent space, where variance indicates the uncertainty in the feature space.
%
%
Quality can be understood as a measure of the utility of a sample for recognition, so samples with higher quality lead to a higher accuracy \cite{[Alonso12a]}.
In this regard, several other methods have been proposed for face quality analysis (see references within \cite{FuDamer22WACV_FIQA_Explainability}), 
but they do not provide spatial interpretations. 
To address this, \cite{FuDamer22WACV_FIQA_Explainability} proposed to explain the decision of face quality algorithms by analyzing its impact on the activations of the deepest convolution layer of a FR CNN.
They found that high-quality images have low activations outside the central face region, while low-quality images exhibit variability across face areas.
%
%
%
LIME was recently applied \cite{Rajpal23WPC_LIME_FR_explainability} to interpret FR networks, but only in closed-set identification using the same training classes. In addition, explainability is provided by visual analysis of just a few images.
%

\section{Explainability Approach}
\label{sec:LIME}



This section presents the adapted LIME method \cite{Ribeiro16LIME} for biometric verification.
The image is first divided into superpixels (Figure~\ref{fig:LIME_method}) to group similar-value pixels.
Then, the image is perturbated by randomly blacking out certain superpixels.
Each perturbed image is processed by the CNN, producing softmax probabilities for the target class to be explained.
Next, a weighted linear classifier is trained to predict softmax scores using superpixel contributions.
Superpixel activity is represented by a binary vector, e.g. if there are three superpixels, a vector [1 0 1] indicates that the second is masked.
Also, weights are assigned based on the similarity of perturbated samples to the original image.
%
%
LIME assumes an approximately linear decision boundary near the sample, giving higher weight to closer perturbed samples.
%
%
%
%
%
%
Linear coefficients represent superpixel importance towards the prediction.
Thus, the pixels of each superpixel are set to the value of the associated coefficient.
Smoothing can be applied using a Gaussian kernel, and output is scaled to [0,1] for heatmap comparability.
%
%
%

To use the model for verification, softmax scores are replaced with cosine similarities between the input image's feature vector from a specific network layer and the feature vectors of perturbed images.
This approach does not require the classification head and enables the explanation of classes beyond the training set.
%
%
%
The intuition is that perturbation of highly relevant areas will result in feature vectors much less similar to the original, in the same way, that the softmax probability of the actual class should be smaller.
Therefore, if removing a superpixel significantly alters the vector, that superpixel is considered highly important.
%
%
%
%
In this work, we use the LIME implementation of \cite{LIMEmatlab}, which we modify as described.
We employ 75 superpixels,
1000 perturbated images,
%
and $\sigma$=4 in the smoothing kernel.
To generate a perturbated image, each superpixel is blacked with a probability of 60\%.

\section{Materials and Methods}
\label{sec:intro}

\subsection{Face Recognition Networks}
\label{sec:CNNs}

We use two backbone architectures: MobileNetv2 \cite{[Sandler18mobilenetv2]} (light) and ResNet50 \cite{[He16]} (large).
They respectively have 53/50 convolutional layers and 3.5M/25.6M parameters.
ResNet introduced residual blocks that bypass intermediate layers, improving gradient propagation and allowing deeper networks without overfitting.
In a residual layer, channel dimensionality is first reduced via 1$\times$1 point-wise filters, 
and then increased again to match the input.
MobileNets employ inverted residuals and depth-wise separable convolutions to reduce parameters and inference time.
Shortcut connections are between thinner layers instead (hence the name 'inverted'), which also results in fewer parameters.
%
%
%
%
The choice of these architectures allows for comparison between a light and heavy network.
%
%
The original models are modified to have an input of 113$\times$113$\times$3 by changing the stride of the first convolutional layer from 2 to 1.
This allows to keep the rest of the network unchanged and to reuse ImageNet as starting model.

\subsection{Data and Biometric Verification Protocol}
\label{sec:data}

The VGGFace2 database \cite{[Cao18vggface2]} is used for training and evaluation (3.31M images of 9,131 celebrities, 363.6 images/person on average).
The images exhibit pose, age, ethnicity, lighting, and background variations.
The training protocol involves 8,631 classes (3.14M images) for training and 500 classes for testing.
For cross-pose experiments, a subset (VGGFace2-Pose) is defined, with 368 subjects from the test set with 10 images per pose (frontal, three-quarter, and profile), resulting in 11,040 images.
To improve recognition performance, we utilize the RetinaFace cleaned set of the MS-Celeb-1M database (MS1M) \cite{[Guo16_MSCeleb1M]}, with 5.1M images of 93.4K identities.
%
%
Although MS1M has a larger image count, it has limited intra-identity variation with an average of 81 images/person.
Following previous research \cite{[Cao18vggface2],[Alonso20SqueezeFacePoseNet]}, we adopt a two-step training approach. We first train the models on MS1M, and then fine-tune them on VGGFace2, which provides greater intra-class diversity. This strategy has demonstrated superior performance compared to training solely on VGGFace2.
%
%

The networks are trained for biometric identification using cross-entropy loss and ImageNet initialization.
We follow the training/evaluation protocol of VGGFace2 \cite{[Cao18vggface2]}.
The bounding boxes of VGGFace2 images are resized, so the shorter side has 129 pixels, and a crop of 113$\times$113 is taken.
%
%
MS1M images are directly at 113$\times$113 (thus, no crop).
During training, horizontal random flip is applied.
The optimizer is SGDM (batch=128, learning rate=0.01, 0.005, 0.001, and 0.0001, decreased when the validation loss plateaus).
Two percent of images per user in the training set are set aside for validation.
MS1M under-represented users ($<$70 images) are removed to ensure at least one validation image per user, resulting in 35,016 users/3.16M images.
%
%
Training is performed using Matlab r2022b, utilizing the pre-trained ImageNet model provided.

Verification experiments are done with VGGFace2-Pose.
Genuine scores are obtained by comparing each image of a user with the remaining images of the same user, excluding symmetric matches. For impostor scores, the 1$^{st}$ image of a user is compared with the 2$^{nd}$ image of the next 100 users.
Different CNN layers are used as feature descriptors to evaluate accuracy across different layers and correlate them with the explanations provided.
To mitigate the impact of pose variation, the descriptor of an image and its horizontally flipped counterpart are averaged
\cite{[Duong19MobiFace]}.
The verification score between two images is obtained by cosine similarity.

\begin{figure}[b]
\centerline{\includegraphics[width=0.4\textwidth]{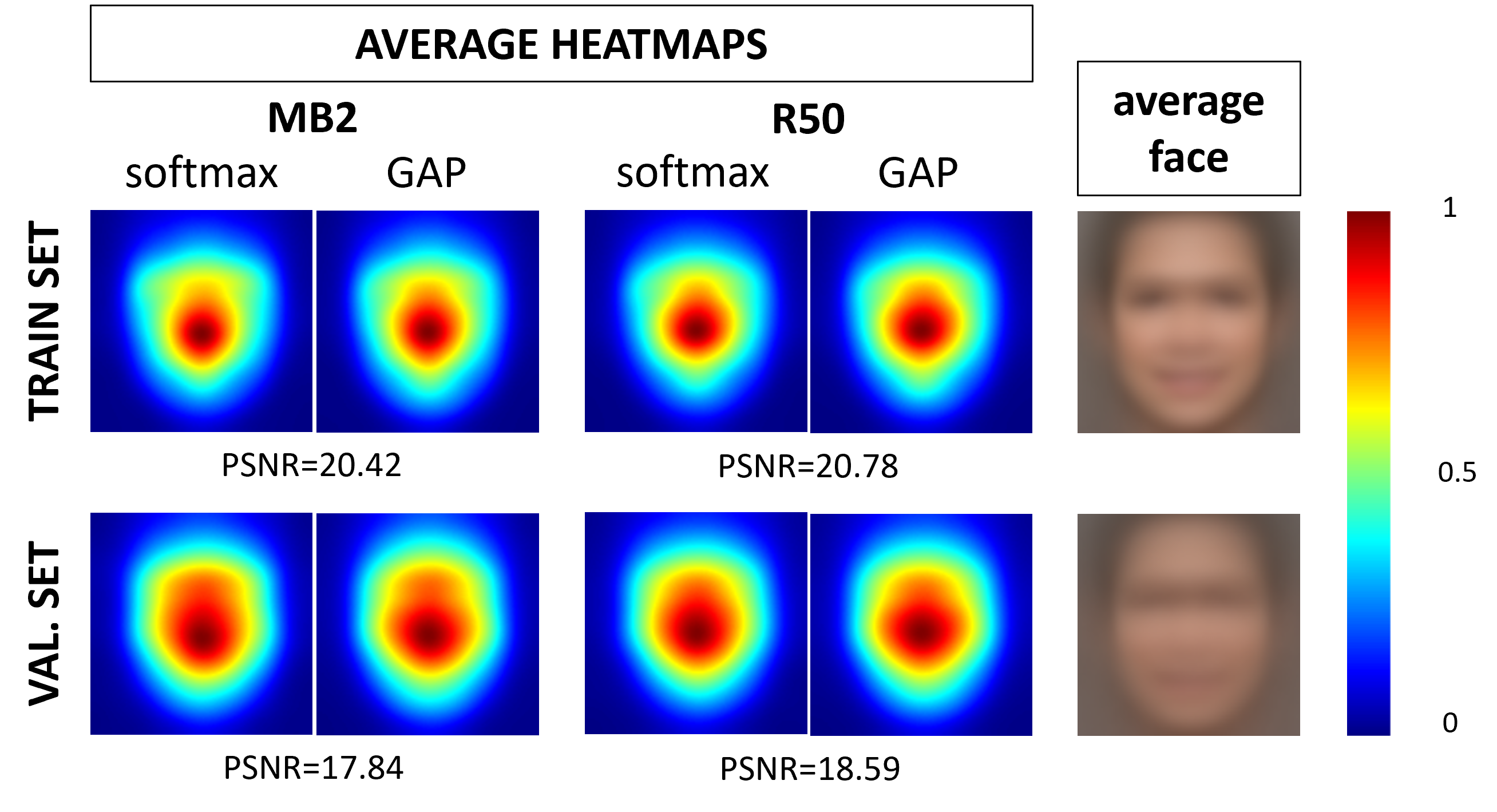}}
\caption{Average heatmap and PSNR per data set, vector source, and CNN.}
\label{fig:avg_heatmaps_train_val}
\end{figure}

\begin{figure}[b]
\centerline{\includegraphics[width=0.3\textwidth]{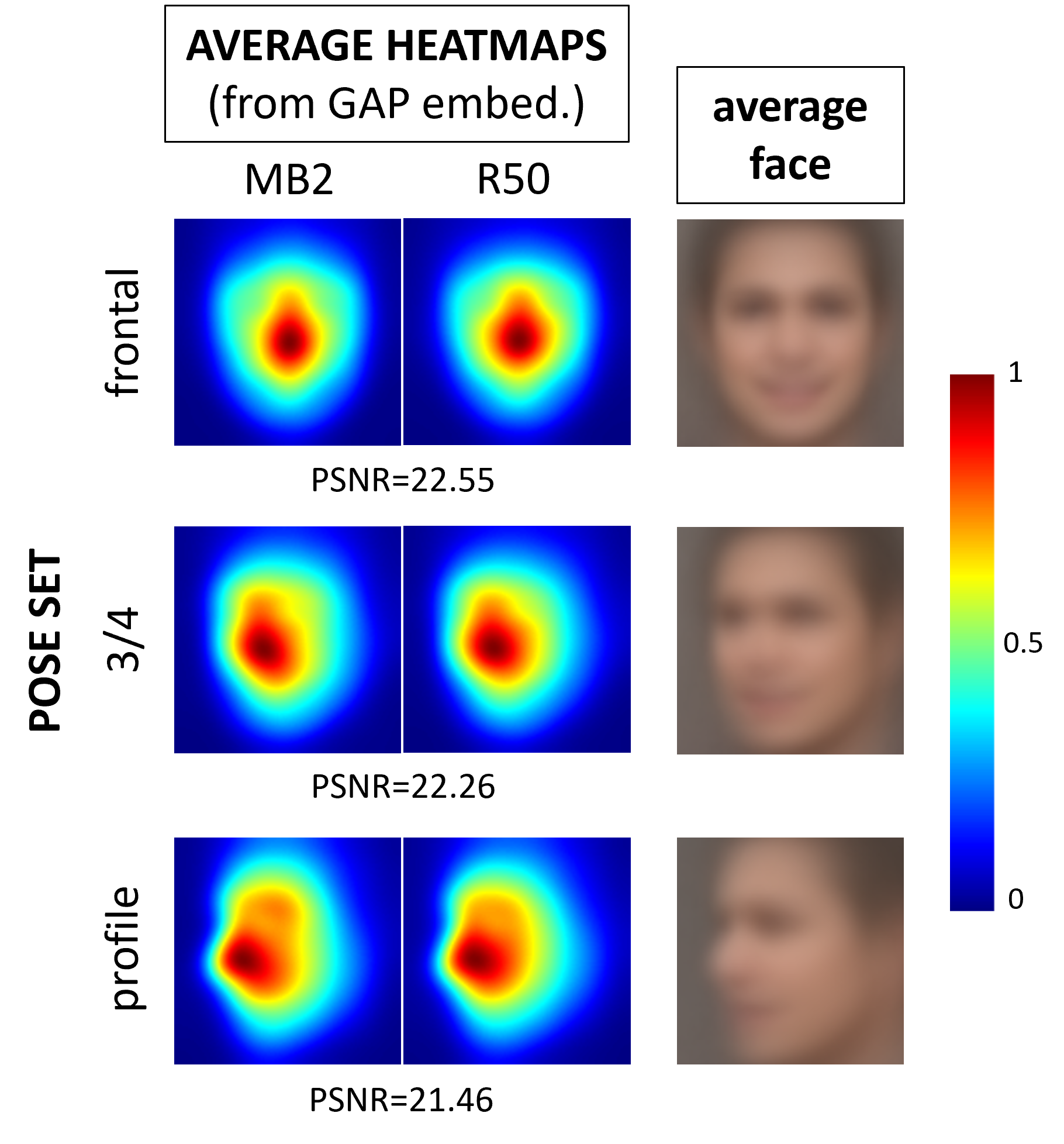}}
\caption{Average heatmap and PSNR per pose set and CNN.}
\label{fig:avg_heatmaps_pose}
\end{figure}

\section{Results}
\label{sec:results}

\subsection{GAP embedding vs. Softmax Heatmaps}
\label{sec:GAPvsSoftmax}

We compare the heatmaps from the embedding before the classification head (our proposal) with the softmax probability (original LIME method).
The embedding (GAP layer, or Global Average Pooling) has a dimensionality of 1280 (MobileNetv2) and 2048 (ResNet50).
Figure~\ref{fig:avg_heatmaps_train_val} shows the average heatmaps of VGGFace2 training and validation sets.
To speed up computations, we only use one training and one validation image per user (8631 training + 8631 validation images).
We also show the average PSNR between the heatmaps from the embedding and the softmax probability.

Focusing on the training set, it can be seen that, on average, the relevant regions in the GAP and softmax heatmaps are the same.
Both networks primarily emphasize the nose and its surroundings, including the area between the eyes.
The eyes and mouth also receive attention, though to a lesser extent.
However, most face regions have an attention score above 0.4, so their impact on recognition is not negligible.
In the validation set, the region of attention is similar, but more importance is given to the eye regions, and the highest-relevance area is larger.
The average face in the validation set appears more blurred in the eye and mouth areas, suggesting a presence of non-frontal images, which contribute to the average heatmap spreading towards the left and right sides.
Regarding the average PSNR between GAP and softmax heatmaps, it is lower in the validation set (by over 2 dB).
%
%
%
Also, ResNet50 exhibits slightly higher PSNR, indicating more consistency in its GAP vs. softmax heatmaps.
%
However, based on the similarity between average heatmaps and PSNR values, we conclude that GAP embeddings, as proposed in Section~\ref{sec:LIME}, are a valid source for heatmap computation without relying on softmax information from training classes.
This makes the suggested approach suitable for scenarios like face verification, where unseen classes are involved.

\subsection{Comparing the Two CNNs}

%
We then analyze (Figure~\ref{fig:avg_heatmaps_pose}) the average heatmaps of each network on the VGGFace2-Pose set.
All 3/4 and profile images have been processed to ensure the person always looks to the left side.
The findings align with the previous section.
The nose and inter-eye regions receive the highest attention.
The importance gradually decreases towards the periphery, although most face regions contribute with an importance score of $>$0.4.
Additionally, the effect of different poses causes the average heatmap to rotate towards the left. In the profile view, where one side of the face is not visible, the networks compensate by placing greater attention to the eye region.

\begin{figure}[t]
\centerline{\includegraphics[width=0.4\textwidth]{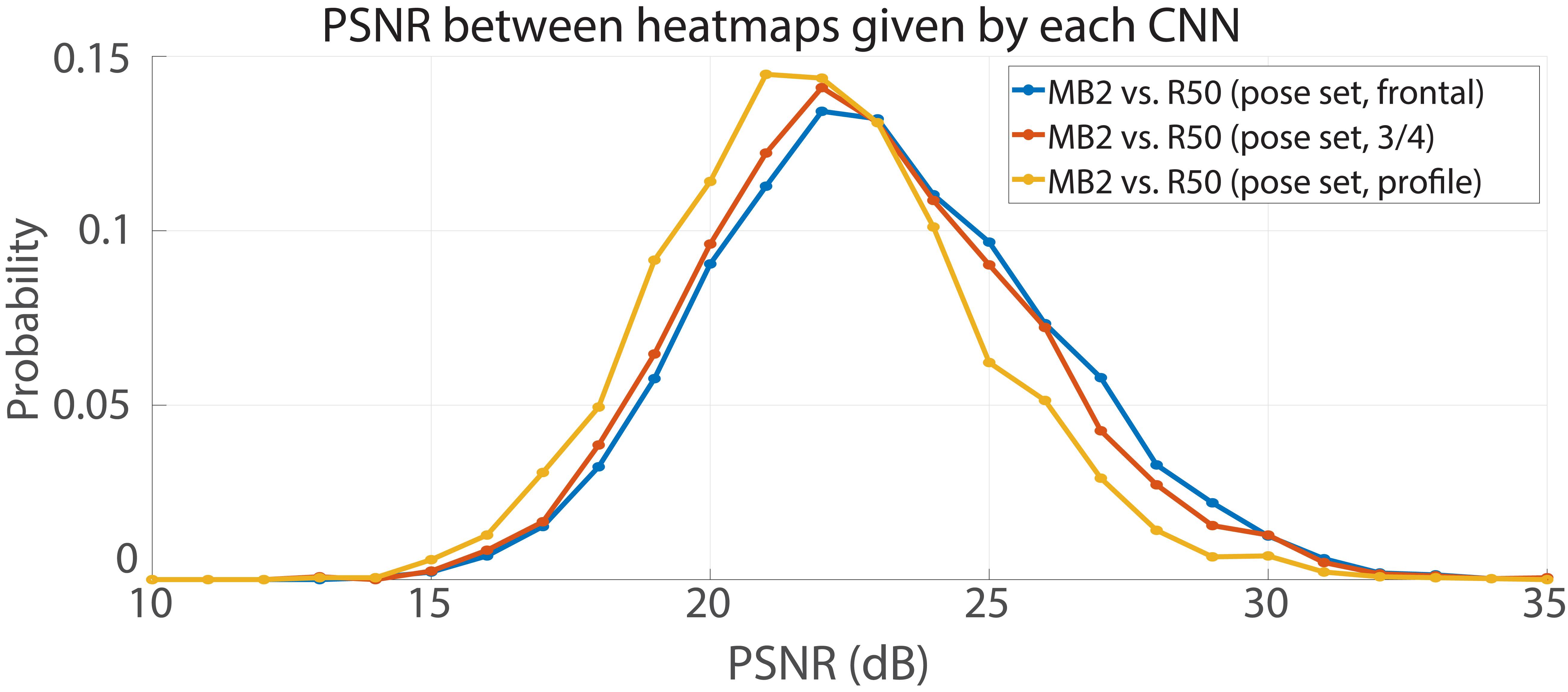}}
\caption{Histogram (per pose) of PSNR between the heatmaps of each CNN.}
\label{fig:PSNR_histogram_between_CNNs_per_pose}
\end{figure}

\begin{figure}[t]
\centerline{\includegraphics[width=0.4\textwidth]{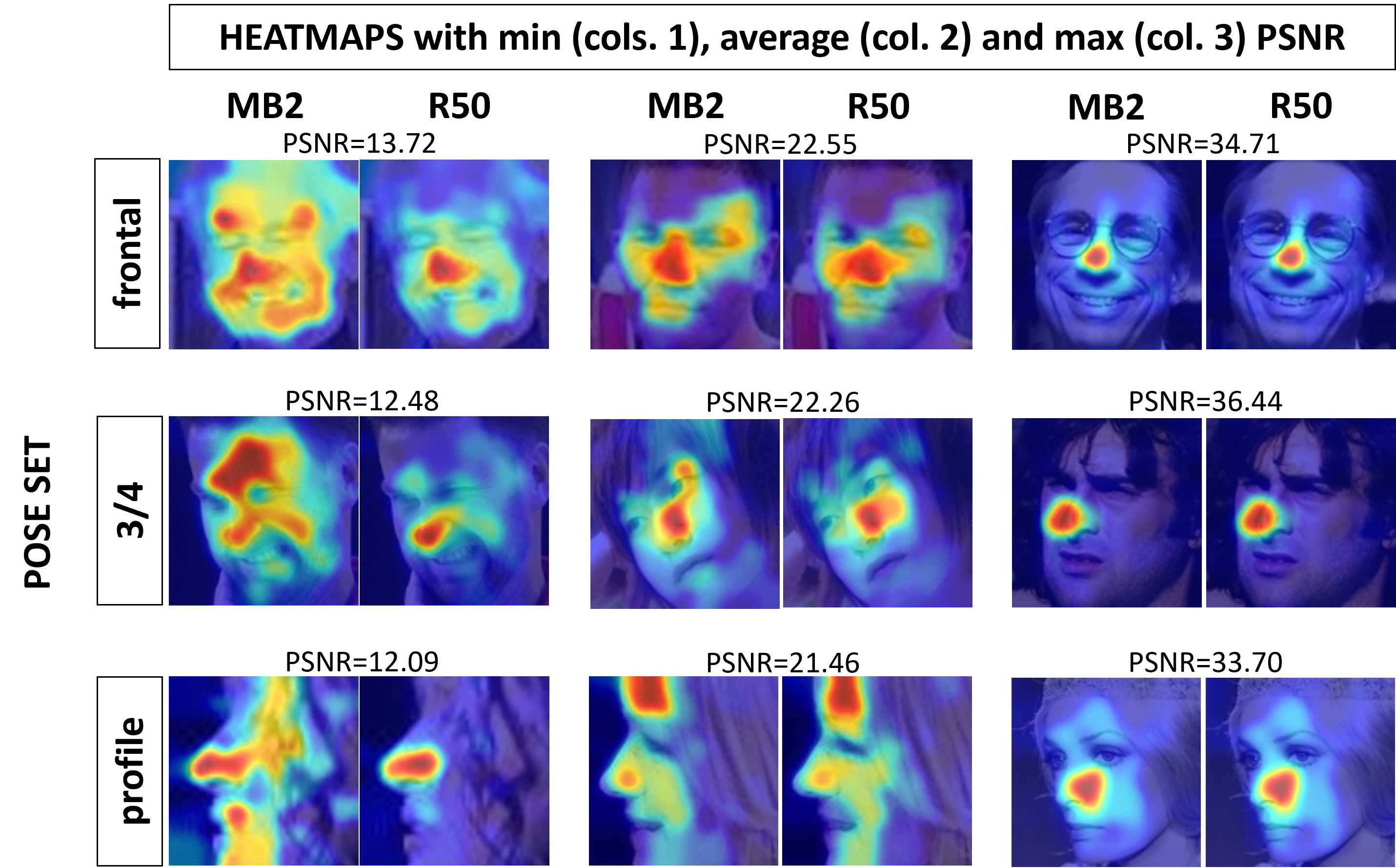}}
\caption{
Individual heatmaps across poses having the maximum, average, and minimum PSNR between the two CNNs.
}
\label{fig:individual_heatmap_per_pose_last_layer}
\end{figure}

While the average heatmaps of the two networks appear to be similar, we also examined their differences at the individual image level.
Figure~\ref{fig:PSNR_histogram_between_CNNs_per_pose} displays the histograms of the PSNR between the heatmaps generated by each CNN on VGGFace2-Pose images.
The PSNR ranges from 14 to 33 dB, indicating that there are variations in the heatmaps for some images.
To investigate this further, we present the heatmaps of individual images with the highest, average, and lowest PSNR in Figure~\ref{fig:individual_heatmap_per_pose_last_layer}.
For the average PSNR (21-22 dB), the heatmaps exhibit visual similarity with minor differences. In the case of the lowest PSNR, both networks highlight common regions of strong influence, such as the nose.
However, there are additional regions emphasized by only one network.
In the instances shown in Figure~\ref{fig:individual_heatmap_per_pose_last_layer}, MobileNetv2 highlights larger areas of the image as important, although these observations are based on only three images and should not be generalized.

\begin{figure}[t]
\centerline{\includegraphics[width=0.4\textwidth]{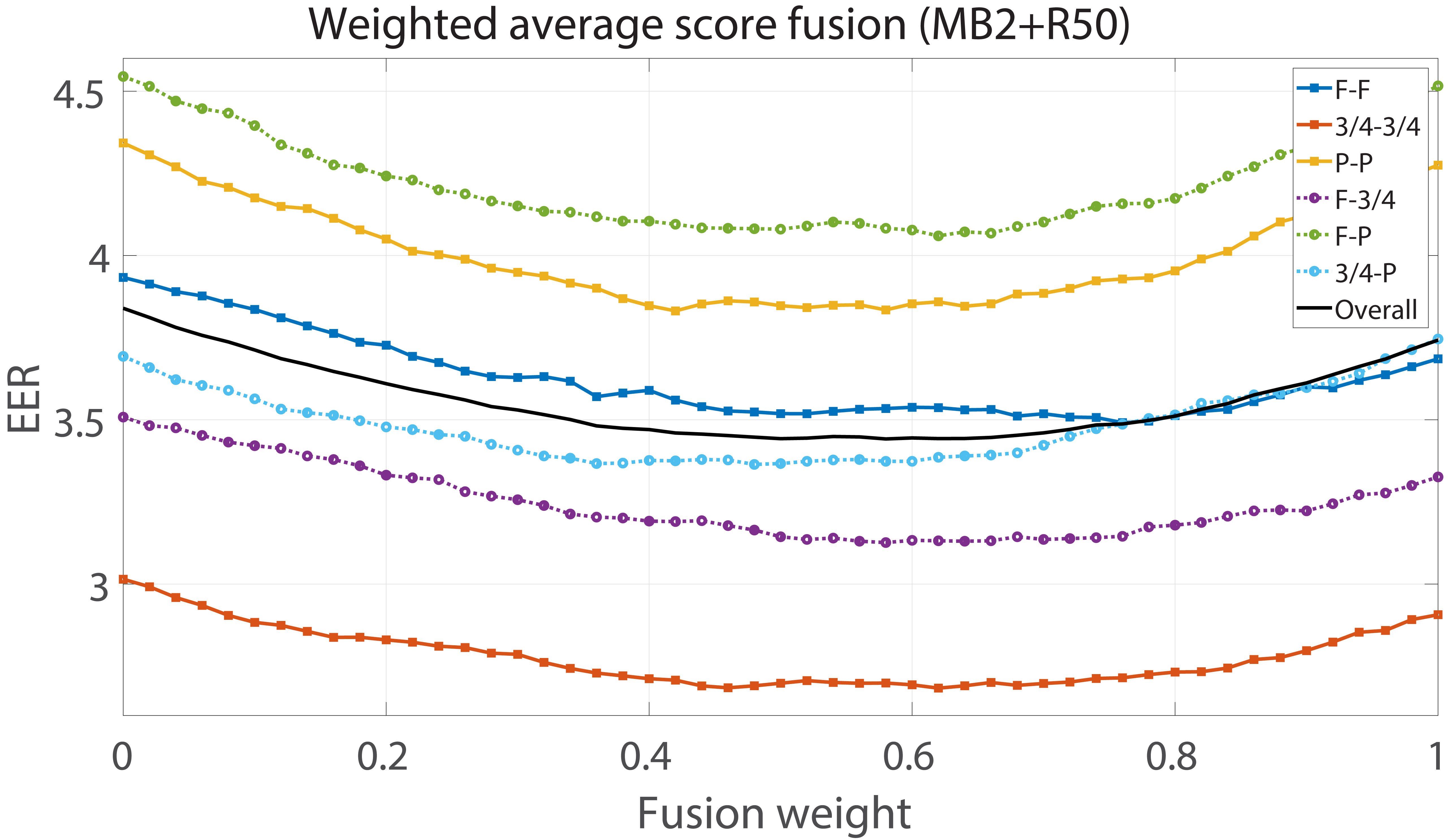}}
\caption{Weighted average score fusion of the two CNNs.}
\label{fig:EER_weighted_fusion}
\end{figure}

\subsection{Network Complementarity}

In the previous subsection (Figure~\ref{fig:PSNR_histogram_between_CNNs_per_pose}), it was noticed that the PSNR between heatmaps of the two CNNs decreases as faces deviate from a frontal view (see the leftward shift in the histograms).
Along with the low PSNR for a portion of the images,
it suggests that the networks might be complementary, even though they generally focus on similar regions. Furthermore, complementarity may vary depending on the pose.

Table~\ref{tab:results-EER} shows the verification accuracy on the VGGFace2-Pose dataset, including same- and cross-pose experiments, as well as overall performance across all poses.
Same-pose comparisons are made using GAP vectors from images of the same pose, while cross-pose experiments involve images with different poses.
The two networks demonstrate similar performance, with differences typically below 0.2\% across different pose cases.
Notably, the most challenging scenarios are observed when the image is only visible from one side (e.g., Profile vs. Profile) or when there is a significant difference between query and test templates (e.g., Frontal vs. Profile).

We then assess the complementarity of the networks by combining their verification scores, denoted as $s_{MB2}$ and $s_{R50}$, through a weighted average approach via $a \times s_{MB2} + (a-1) \times s_{R50}$ 
($a\in [0,1]$).
%
%
Figure~\ref{fig:EER_weighted_fusion} shows the results for different values of the weight $a$.
Notably, the fusion enhances performance across all pose cases, with the optimal performance achieved when both networks are assigned a roughly equal weight ($a$ between 0.4 and 0.6).
We select the case with the highest overall accuracy ($a=0.58$) and provide its exact EER values in the last row of Table~\ref{tab:results-EER}.
Apart from the mentioned improvement, the biggest reduction in EER (by about 10\%) is observed in the mentioned challenging cases of Profile vs. Profile and Frontal vs. Profile.
This aligns with the earlier finding that the PSNR between heatmaps of the two CNNs tends to decrease as faces deviate from a frontal view.
Consequently, the greatest complementarity is observed under challenging cases involving non-frontal (i.e., profile) images.

\begin{table}[t]

\caption{Verification results on the VGGFace2-Pose database (EER \%). F=Frontal View. 3/4= Three-Quarter. P=Profile.}

\centering

\begin{adjustbox}{max width=\columnwidth}

\begin{tabular}{|c|ccc|ccc|c|}

\cline{2-8}

\multicolumn{1}{c}{}  & \multicolumn{3}{|c|}{\textbf{Same-Pose}} &  \multicolumn{3}{c|}{\textbf{Cross-Pose}} & \textbf{Over-}  \\  \cline{2-7}

\multicolumn{1}{c|}{\textbf{Network}}

& \textbf{F-F} & \textbf{3/4-3/4} & \textbf{P-P}

& \textbf{F-3/4} & \textbf{F-P}  & \textbf{3/4-P}  & \textbf{all}

\\ \hline

\textbf{MB2}  &

3.69 &	2.91 &	4.27 &	3.33 &	4.52 &	3.75 &  3.75

\\ \hline

\textbf{R50} &

3.93 &	3.01 &	4.34 &	3.51 &	4.54 &	3.69 & 3.84

\\ \hhline{========}

\textbf{combined} &


3.53 &	2.70 &	3.83 &	3.13 &	4.08 &	3.37 &	3.44 \\

\textbf{(best)} & (-4.1\%) &   (-7.2\%) &  (-10.3\%) &   (-6.0\%) &   (-9.6\%) &   (-8.6\%) &  (-8.0\%)

\\ \hline


\end{tabular}


\end{adjustbox}

\label{tab:results-EER}

\end{table}

\begin{figure}[t]
\centerline{\includegraphics[width=0.48\textwidth]{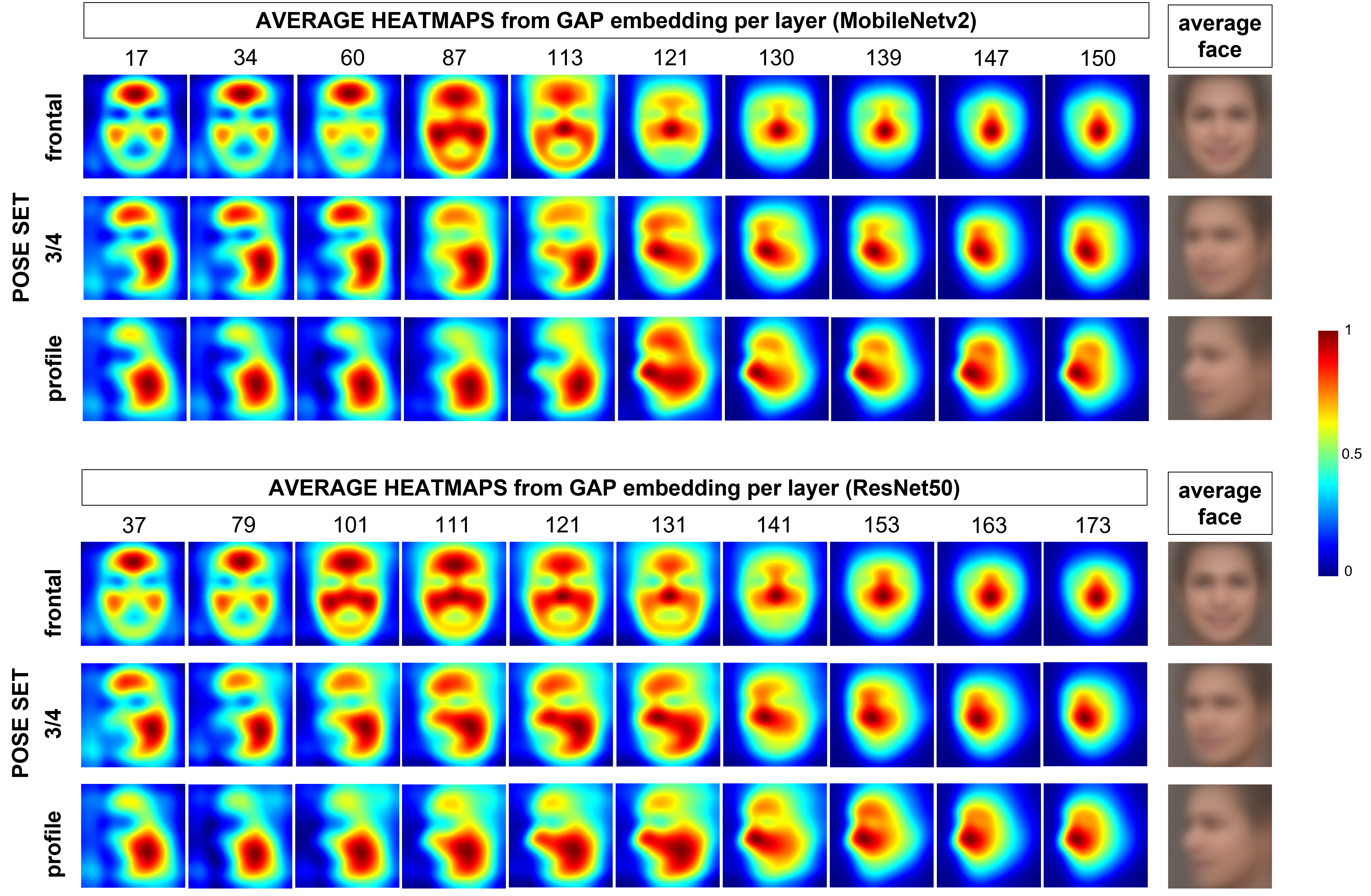}}
\caption{Average heatmaps across different layers of the networks.}
\label{fig:avg_heatmap_per_pose_per_layer}
\end{figure}

\begin{figure}[t]
\centerline{\includegraphics[width=0.4\textwidth]{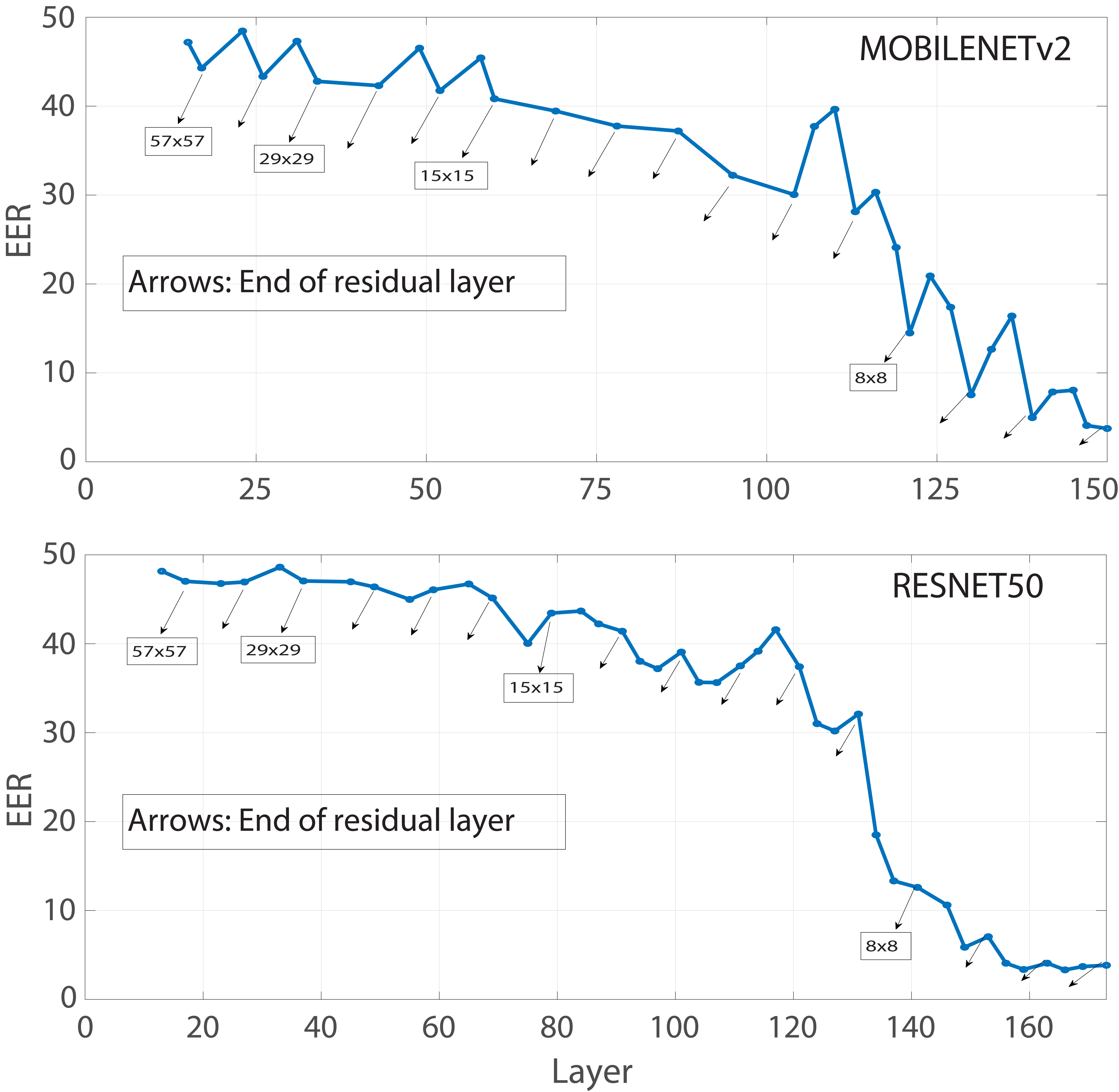}}
\caption{EER across different layers of the networks.}
\label{fig:EER_per_layer}
\end{figure}

\subsection{Learning Across Different Layers}

We now examine the evolution of heatmaps across different layers in Figure~\ref{fig:avg_heatmap_per_pose_per_layer}.
Additionally, we present the Equal Error Rate (EER) per layer in Figure~\ref{fig:EER_per_layer}.
To generate heatmaps and verification scores, we utilize the embedding provided by the respective intermediate layer.
The layer number correspond to the Matlab r2022b models.
MobileNetv2 consists of 16 inverted residual blocks, and GAP embedding occurs after layer 150. The residual blocks end at layers 17, 26, 34, 43, 52, 60, 69, 78, 87, 95, 104, 113, 121, 130, 139, and 147 (indicated by arrows in Figure~\ref{fig:EER_per_layer}). Downsampling occurs after the 1$^{st}$, 3$^{rd}$, 6$^{th}$, and 13$^{th}$ block.
On the other hand, ResNet50 comprises 16 residual blocks, with the GAP embedding situated after layer 173. The residual blocks terminate at layers 17, 27, 37, 49, 59, 69, 79, 91, 101, 111, 121, 131, 141, 153, and 173. Downsampling happens after the 1$^{st}$, 4$^{th}$, 8$^{th}$, and 14$^{th}$ block.

\begin{figure*}[htbp]
\centerline{\includegraphics[width=0.95\textwidth]{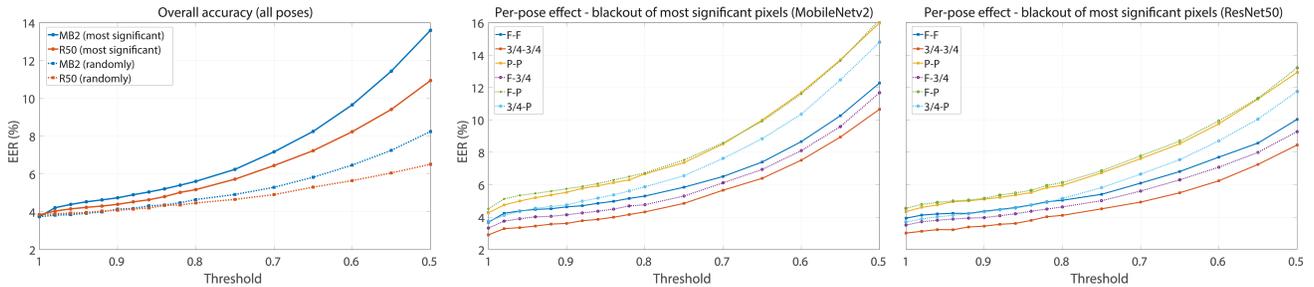}}

\caption{Effect on the EER of eliminating the most significant pixels.}
\label{fig:blackout_EER}
\end{figure*}

\begin{figure}[htbp]
\centerline{\includegraphics[width=0.45\textwidth]{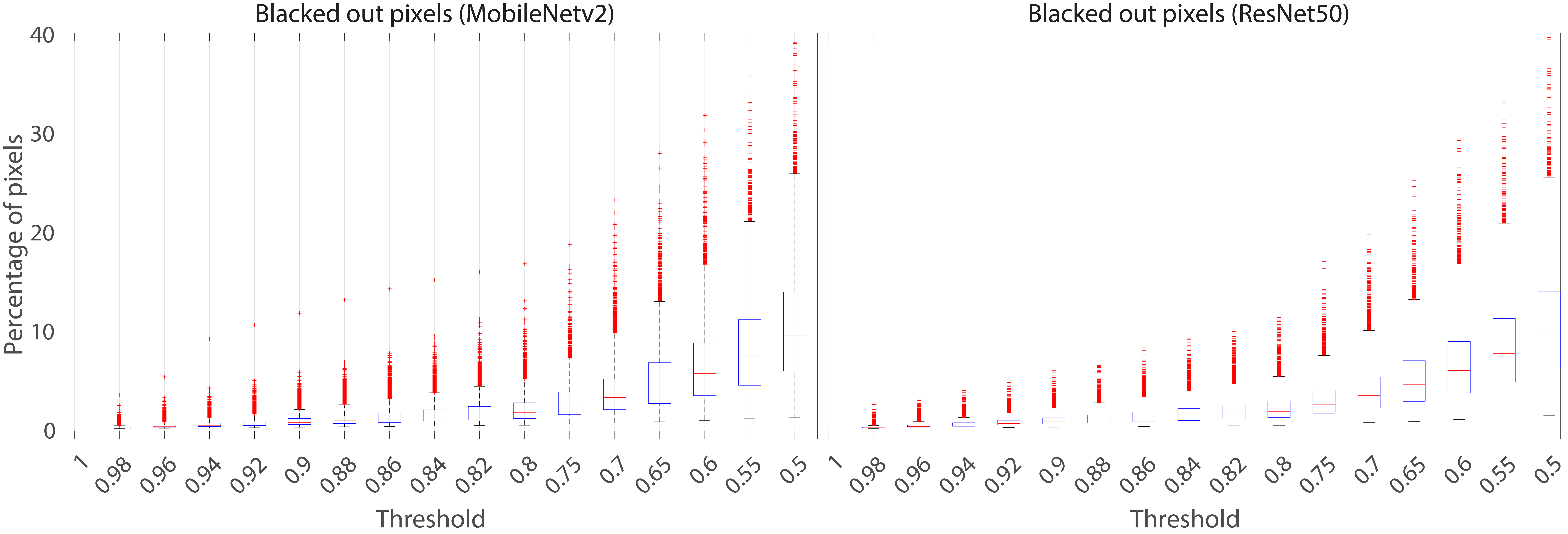}}

\caption{Percentage of most significant pixels removed per threshold.}
\label{fig:blackout_amount_pixels}
\end{figure}

\begin{figure}[htbp]
\centerline{\includegraphics[width=0.45\textwidth]{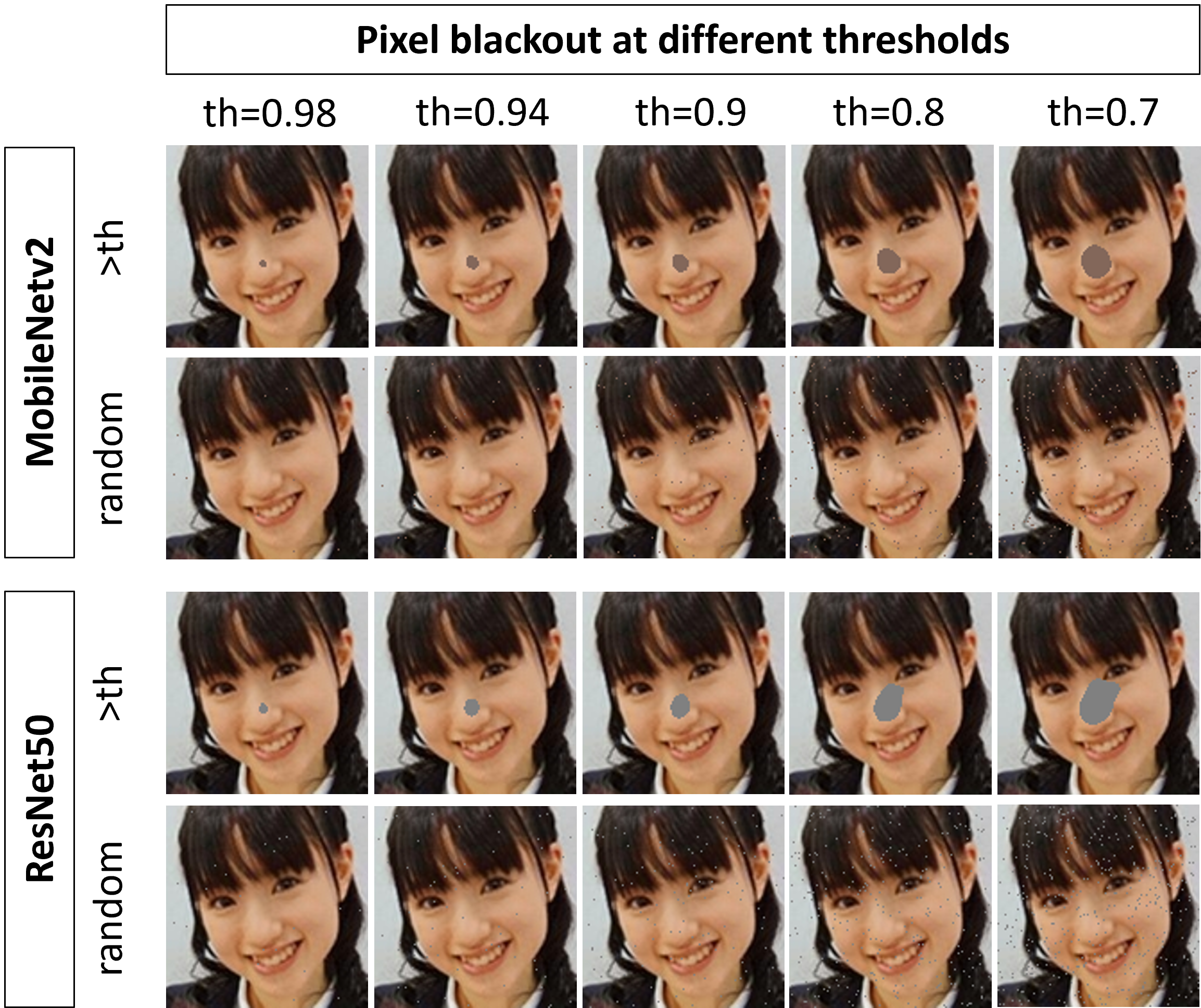}}
\caption{Example image with the most significant pixels eliminated.}
\label{fig:blackout_examples}
\end{figure}

In Figure~\ref{fig:avg_heatmap_per_pose_per_layer}, it can be observed that both networks exhibit similar evolution patterns across layers.
In the initial layers, the emphasis is primarily on the forehead and cheek skin. In profile poses, where the forehead is less visible, the highest importance shifts to the cheek area. As we progress to mid layers, larger portions of the forehead and cheek skin are considered the most important.
Eventually, the networks converge to prioritize the regions mentioned in Section~\ref{sec:GAPvsSoftmax}, namely the nose and its surrounding areas, including the inter-eye region.
Interestingly, the eye and mouth areas receive relatively low importance in the initial layers. However, in the mid layers, these areas gain some importance (indicated by a yellow-ish color, approximately 0.6), which is then slightly reduced in the final layers (indicated by a green-ish color, around 0.5).

In terms of verification accuracy (Figure~\ref{fig:EER_per_layer}), both networks initially exhibit poor performance, 
%
and the EER decreases gradually but slowly. However, when approximately 75\% of the network has been traversed, the reduction becomes more rapid.
An interesting observation with MobileNetv2 is that the EER across residual layers first increases and then decreases towards the end of the layer, as indicated by the oscillations between the arrows.
On the other hand, ResNet50 shows the opposite effect, with the EER being lower in the middle of the residual layers. 
This behavior can be attributed to the design principles of the two networks.
MobileNetv2 follows a structure where the number of channels is low at the beginning and end of a residual block but high in the middle. In contrast, ResNet50 follows the opposite principle, with higher dimensionality at the beginning and end, and lower dimensionality in the middle.
Therefore, a common trend observed in both networks is that the lowest EER in a block is typically achieved by vectors from low-dimensional layers.

\subsection{Eliminating the Most Significant Pixels}

To validate our explainability approach further, we conduct verification experiments by iteratively removing pixels with the highest importance.
Figure~\ref{fig:blackout_EER} shows the effect on the EER as pixels above a certain threshold are eliminated.
Specifically, they are set to the mean normalization value of the input layer, effectively becoming zero when the network normalizes the image.
In Figure~\ref{fig:blackout_amount_pixels}, boxplots depict the percentage of pixels removed at each threshold. Furthermore, Figure~\ref{fig:blackout_examples} presents an example of input images.
We compare this approach to randomly removing the same number of pixels.

In Figure~\ref{fig:blackout_EER}, it is evident that eliminating the most significant pixels leads to a quicker decrease in the EER compared to randomly removing the same number of pixels.
This suggests that the pixels selected using the explainability method hold more importance than random pixels.
Additionally, the left plot shows that MobileNetv2 is more adversely affected by pixel elimination than ResNet50, indicating that the latter is more resilient to removing significant facial areas.
When examining verification accuracy for different poses (center and right plots), it becomes apparent that combinations involving profile images (green, yellow, and light blue) are highly sensitive to the elimination of significant pixels.
This sensitivity is reflected in the higher increase in EER as more pixels are removed.
The effect is specially pronounced in the Three-quarter vs. Profile case (light blue curve).
These results highlight the criticality of preserving significant pixels, especially when the pose deviates from a frontal view.

From Figure~\ref{fig:blackout_amount_pixels}, it is evident that as the threshold decreases, the range between the 25$^{th}$-75$^{th}$ percentiles and the whiskers increases. This indicates that the percentage of pixels above the threshold varies significantly among images.
%
%
Particularly noteworthy is that at 0.5, the average percentage of removed pixels is below 10\%. However, the EER for both networks has multiplied by $\sim$3.
This emphasizes the removed pixels' importance and the networks' high sensitivity to their elimination.
When no pixels are removed, the EER is 3.75/3.84\% for MB2/R50 (Table~\ref{tab:results-EER}).
However, at a threshold of 0.8, where only 1.6-1.8\% of pixels are removed on average, the overall EER increases to 5.61/5.17\% for MB2/R50.
This is a whooping EER increase of about 35-50\% after removing such a small proportion of pixels.
%

\section{Conclusions}
\label{sec:conclusions}

Vision-based biometrics, like many other vision tasks, currently relies heavily on deep learning models \cite{[Sundararajan18-DLbiometrics]}.
However, the lack of interpretability of these models has garnered significant attention due to their black box nature when it comes to decision-making \cite{Jain22PAMI_biometrics_trust_verify}.
In this study, we apply the LIME (Local Interpretable Model-agnostic Explanation) method \cite{Ribeiro16LIME} to identify pixels that are most relevant for recognition.
Our focus is biometric verification, with face recognition (FR) as case study, although the method is applicable to any modality.
LIME is designed initially to explain the classes used for training by utilizing the softmax probabilities of the classification head.
However, biometric verification is usually conducted on different classes (identities) that those used for training by using embeddings from a layer before the classification head.
Therefore, we adapt LIME to utilize these embeddings instead.
Experiments are conducted with two CNNs based on MobileNetv2 and ResNet50 trained for face recognition over the MS-Celeb-1M \cite{[Guo16_MSCeleb1M]} and VGGFace2 \cite{[Cao18vggface2]} sets.

We first validated our adaptation of LIME by comparing the average maps obtained with softmax probabilities and layer embeddings over the training classes.
On average, these maps are visually similar, and both networks focus on the central (nose) area of the face, with attention diminishing toward the periphery.
We then conducted experiments on VGGFace2-Pose, a test set of unseen identities with three different poses.
At the image level, we observed that the networks provide different heatmaps for a portion of the images, particularly as the images deviate from a frontal view. This suggests that the networks may be complementary, focusing on different regions.
We confirmed this by combining them via score fusion, resulting in improved performance, especially for non-frontal images where the heatmaps differ the most.
%

%
%

We then applied the proposed method to visualize the heatmaps of the networks across different layers.
We observed that the networks vary their attention to different regions, such as the forehead and cheek at low initial layers and the nose and surrounding parts at late layers.
It would be interesting to explore mechanisms to leverage information from layers that employ knowledge from different parts of the face.
Additionally, we noticed that the networks put modest attention to the eyes and mouth, despite these regions being regarded as highly discriminative in the literature \cite{[Alonso16]}.
Therefore, another direction for future research would be to consider mechanisms to increase attention to these regions \cite{HappyDantcheva21IVS_Expression_Part_Faces}.

Lastly, we demonstrated that eliminating significant pixels of the face marked by the LIME method significantly impacts verification accuracy compared to removing the same amount of pixels from random regions.
By removing less than 2\% of the pixels, the EER increases by 35-50\% 
%
%
However, one of the networks showed higher robustness to such elimination. 
This opens up avenues for investigating the effect on face occlusions.
Networks with attention to different face regions \cite{Yin19_ICCV_Interpretable_FR} or networks affected differently by eliminating such areas could help minimize the impact of occlusions through appropriate network combinations.
We also observed that eliminating significant pixels is more critical for non-frontal images.
It would be interesting to assess if the training database contains an adequate amount of non-frontal data, as this may contribute to the phenomenon.
However, non-frontal face recognition is a more challenging problem due to occlusions in parts of the face, as seen by the highest EER.
In such cases, we noticed that the networks try to compensate by focusing more on the eye region.
A future direction could involve employing networks explicitly trained for different poses to address this issue.

Another avenue for future work relates to the findings in \cite{FuDamer22WACV_FIQA_Explainability}, which indicated that high-quality images tend to have low activations outside the central face region, while low-quality images exhibit higher variability.
We observed similar behavior in terms of heatmap agreement between the two networks (Figure~\ref{fig:individual_heatmap_per_pose_last_layer}).
Therefore, we speculate that low-quality images may increase disagreement (lower PSNR) between the networks' heatmaps. This suggests the possibility of utilizing our explainability method as a tool to assess image quality.
Furthermore, we anticipate that the networks would exhibit greater complementarity with low-quality or high-disagreement images since the networks focus on different parts of the face.

\small

\section*{Acknowledgment}

This work was partly done while F. A.-F. was a visiting researcher at the University of the Balearic Islands.
Authors F. A.-F., K. H.-D., P. T., and J. B. thank the
Swedish Research Council (VR) and the Swedish Innovation Agency (VINNOVA) for funding their research.
Author J. M. B. would like to
thank the project EXPLAINING - "Project EXPLainable Artificial INtelligence systems for health and well-beING", under Spanish national projects funding (PID2019-104829RA-I00/AEI/10.13039/501100011033).
We gratefully acknowledge
the support of NVIDIA Corporation with the donation of the Titan V GPU used for this research.
The data handling in Sweden was enabled by the National Academic Infrastructure for Supercomputing in Sweden (NAISS).


\bibliographystyle{IEEEtran}


\end{document}